\documentclass{article}

\usepackage{subcaption}
\usepackage{graphicx}
\usepackage[preprint]{neurips_2023}

\usepackage[utf8]{inputenc} 
\usepackage[T1]{fontenc}    
\usepackage{hyperref}       
\usepackage{url}            
\usepackage{booktabs}       
\usepackage{amsfonts}       
\usepackage{nicefrac}       
\usepackage{microtype}      
\usepackage{xcolor}         
\usepackage{multirow}
\usepackage{amsmath}

\newcommand{\grad}{\texttt{grad}}

\newcommand{\sac}{\texttt{clip-w}}
\newcommand{\rac}{\texttt{clip-r}}
\newcommand{\hc}{\texttt{human-crop}}

\usepackage[sectionbib]{bibunits}
\defaultbibliographystyle{nips}
\defaultbibliography{cite}

\title{On the Difficulty of Answering Fine-Detail Questions in Open-Ended Visual Question Answering}
\title{Capturing Fine Details in Open-Ended Visual Question Answering via Cropping}
\title{Answering Fine-Detail Questions via Visual Cropping}
\title{Enhancing Vision Question Answering of BLIP-Family Models via Visual Cropping}
\title{Enhancing Fine-Detail Question Answering of BLIP-Family Models via Visual Cropping}
\title{Using Visual Cropping to Enhance Fine-Detail Question Answering of BLIP-Family Models}

\author{%
  Jiarui Zhang \\
  Information Sciences Institute\\
  University of Southern California, USA \\
  \texttt{jrzhang@isi.edu} \\
  \And
  Mahyar Khayatkhoei \\
  Information Sciences Institute \\
  University of Southern California, USA \\
  \texttt{mkhayat@isi.edu} \\
  \AND
  Prateek Chhikara \\
  Information Sciences Institute \\
  University of Southern California, USA \\
  \texttt{pchhikar@isi.edu} \\
  \And
  Filip Ilievski \\
  Information Sciences Institute \\
  University of Southern California, USA \\
  \texttt{ilievski@isi.edu} \\
}

\begin{document}

\maketitle

\begin{abstract}
Visual Question Answering is a challenging task, as it requires seamless interaction between perceptual, linguistic, and background knowledge systems. While the recent progress of visual and natural language models like BLIP has led to improved performance on this task, we lack understanding of the ability of such models to perform on different kinds of questions and reasoning types. As our initial analysis of BLIP-family models revealed difficulty with answering fine-detail questions, we investigate the following question: \textit{Can visual cropping be employed to improve the performance of state-of-the-art visual question answering models on fine-detail questions?}
Given the recent success of the BLIP-family models, we study a zero-shot and a fine-tuned BLIP model. We define three controlled subsets of the popular VQA-v2 benchmark to measure whether cropping can help model performance. Besides human cropping, we devise two automatic cropping strategies based on multi-modal embedding by CLIP and BLIP visual QA model gradients.
Our experiments demonstrate that the performance of BLIP model variants can be significantly improved through human cropping, and automatic cropping methods can produce comparable benefits. A deeper dive into our findings indicates that the performance enhancement is more pronounced in zero-shot models than in fine-tuned models and more salient with smaller bounding boxes than larger ones.
We perform case studies to connect quantitative differences with qualitative observations across question types and datasets.
Finally, we see that the cropping enhancement is robust, as we gain an improvement of 4.59\% (absolute) in the general VQA-random task by simply inputting a concatenation of the original and gradient-based cropped images.
We make our code available to facilitate further innovation on visual cropping methods for question answering.

\end{abstract}

\section{Introduction}

Visual Question Answering, or visual QA, is a challenging task that requires effective integration between perceptual, linguistic, and background knowledge~\citep{schwenk2022okvqa}. Achieving robust performance on visual QA tasks is essential to support downstream tasks requiring textual and visual understanding, including semantic scene search~\citep{vo2019reading}, summarization~\citep{rafiq2020scene}, and captioning~\citep{stefanini2022show}. This can benefit applications, such as intelligent traffic monitoring~\citep{xu2021sutd}, and medical diagnosis~\citep{ren2020cgmvqa}.
Given its relevance and challenging nature, visual QA has recently been addressed in open-ended zero- and few-shot settings.

\begin{figure}[h]
    \centering
    \includegraphics[width=0.95\textwidth]{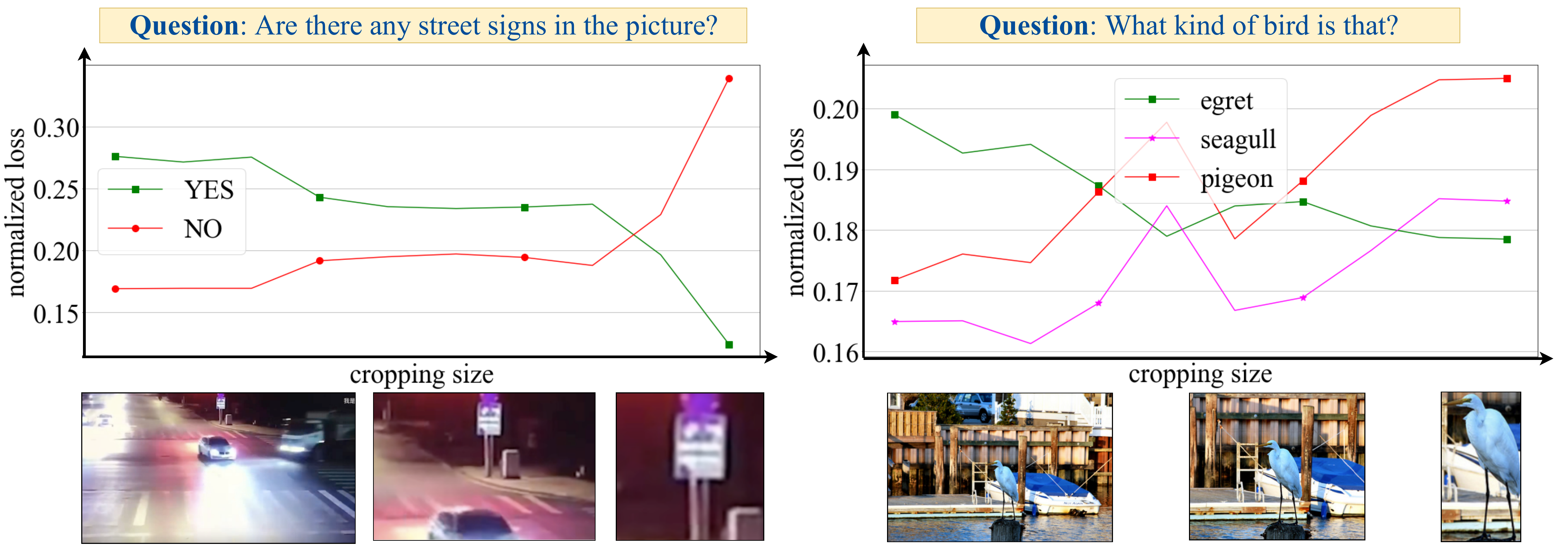}
    \caption{Effect of cropping on visual QA (left: zero-shot and right: fine-tuned) tasks using human cropping (\textit{x-axis} represents cropping size, left-end represents the original image and right-end represents the cropped area of interest). Note that lower loss indicates higher answer confidence.}
    \label{fig:crop_fig}
\end{figure}

State-of-the-art models for open-ended visual QA, like BLIP2~\citep{li2023blip}, typically combine Large Language Models (LLMs) and foundational image embeddings~\citep{brown2020language}. Curiously, while object detection and visual localization have been prevalent tasks \citep{yolo, pascal}, models like BLIP2 work in an end-to-end fashion and do not explicitly include localization components, which raises the question of the mechanism by which they associate the question elements with objects in the image. For instance, given a question like \textit{Are there any street signs in the picture?} (\autoref{fig:crop_fig}), it is unclear whether BLIP-family models focus consistently on the area where the potential street sign is located and whether that focus is clearly leveraged in their QA procedure. Our initial analysis of BLIP2 predictions revealed a recurring theme of mistakes on questions about fine details in images, such as reading text (e.g., letters on the side of an airplane), recognizing whether an object with a specific attribute (e.g., a white dog) exists, or counting objects of a particular type in an image (e.g., slices of pizza). While BLIP-family models have an attention mechanism that, in theory, should localize the important image segment that can help answer the question, we hypothesize that the question-answering module is not consistently attending to the salient part of the image.



Inspired by these insights, we investigate the following question: \textit{Can visual cropping be employed to improve the performance of the BLIP-family visual QA models on fine-detail questions?} 
We illustrate our intuition behind visual cropping with two examples in \autoref{fig:crop_fig}.
While the original BLIP2 method is unable to detect traffic signs in Figure \ref{fig:crop_fig} (left) and identify the bird type in Figure \ref{fig:crop_fig} (right), precise cropping of the image gradually improves its ability to answer both of these questions correctly, switching to \textit{yes} for the first question and to \textit{egret} for the second. 
To provide a systematic insight into our research question, we define three meaningful subsets from the popular VQA-v2 benchmark~\citep{goyal2017making}: \texttt{\textbf{1)}} \textit{VQA-text} contains reading questions, i.e., questions whose answers can be read out from a particular part of the image; \texttt{\textbf{2)}}  \textit{VQA-hard} is selected from errors made both by the BLIP2 zero-shot model and VQA-v2 fine-tuned BLIP model; and \texttt{\textbf{3)}}  \textit{VQA-random} is a random subset that enables us to study the impact of cropping on natural data distribution.
Besides human cropping (\hc), we experiment with novel automatic cropping methods based on the BLIP model gradient (\grad) and on multi-modal embedding by CLIP~\citep{clip} (\sac~and \rac). 
Our experiments demonstrate that the performance of BLIP model variants can be significantly improved through human cropping and that automatic cropping methods can often produce similar benefits. We observe that zero-shot models display more significant improvements, but fine-tuned BLIP models also exhibit enhanced performance. These results indicate that fine-tuning can improve the overall in-domain performance but it does not fully address the issue of insufficient attention to image details. Our findings also indicate that cropping enhancement is robust as it yields an improvement over the baseline on the general VQA-random task.

In summary, our paper makes the following contributions: \texttt{1)} We address a fundamental weakness of visual QA models being unable to reason about non-salient visual elements, by \textbf{measuring the effect of human cropping} on zero-shot and fine-tuned models. \texttt{2)} We devise \textbf{two novel automatic cropping strategies} that leverage multi-modal encoders and vision QA model gradients to approximate human cropping. \texttt{3)} We investigate the value of manual and automatic cropping strategies \textbf{by controlled experiments with zero-shot and fine-tuned models on three new subsets of VQA-v2}. \texttt{4)} We perform \textbf{further analysis} to understand the cropping impact on data partitions, and \textbf{case studies} to connect quantitative differences with qualitative observations across question types and datasets.

\section{Related Work}

\textbf{Vision Language Pre-training (VLP)} models have recently shown promising results across vision-language generation and understanding tasks such as image captioning, visual question answering, and visual grounding \citep{hossain2019comprehensive}. \cite{huang2023language} proposed a multimodal language model that generates text in an auto-regressive manner, perceiving modalities, following instructions, learning in context, and producing outputs. \cite{nguyen2022grit} proposed a Transformer-based architecture for image captioning that integrates region and grid features using Swin Transformer \citep{liu2021swin} and DETR-based detector \citep{carion2020end}.
\cite{zhou2020unified} developed a VLP model with a shared transformer network for encoding and decoding. 
Our approach is orthogonal to these works by focusing on the problem of fine-detail questions in VQA-v2 and investigating the efficacy of visual cropping in improving the performance of state-of-the-art visual QA models. While we test our approach with BLIP-family models, it could be extended to VLP models in the future.

\textbf{Visual QA}
is an emerging research area at the intersection of computer vision and natural language processing, aimed at developing AI models that can answer questions about visual content. \cite{salaberria2023image} proposed a text-only approach for visio-linguistic tasks based on automatic captioning and pre-trained language models, demonstrating improved performance in knowledge-intensive tasks such as OK-VQA and outperforming comparable multimodal models. \cite{alberti2019fusion} introduced Bounding Boxes in Text Transformer (B2T2) architecture to improve visual QA. \cite{garderes2020conceptbert} devised a concept-aware algorithm to answer visual QA questions requiring common sense or basic factual knowledge from external structured content. 
Visual QA models achieve remarkable results but rely on abundant labeled training data. Zero-shot visual QA has gained attention, focusing on answering questions about unseen objects or scenes without task-specific fine-tuning. \cite{pfeiffer2022xgqa} presented the first benchmark for cross-lingual visual QA, with seven target languages. 
\citet{yang2022zero} introduced zero-shot VideoQA using frozen bidirectional language models, achieving state-of-the-art results on benchmark datasets. 
\cite{blip} proposed BLIP, a flexible and transferable model, addressing challenges in understanding and generation-based tasks faced by existing pre-trained models. In follow-up work,
\cite{li2023blip} proposed BLIP2, a generic and efficient pre-training strategy leveraging off-the-shelf frozen pre-trained image encoders and frozen large language models to bootstrap vision-language-language pre-training. 
We take BLIP and BLIP2 as state-of-the-art generalizable models, investigate their ability to answer diverse fine-detail questions, and propose to enhance their ability by novel cropping components.

\textbf{Image Localization}
Accurately localizing objects in images has been a longstanding challenge, driving extensive research into new algorithms and techniques. \cite{clip_localization1} proposed leveraging CLIP for phrase localization without human annotations, showing improved performance over existing no-training methods in zero-shot phrase localization. \cite{kirillov2023segment} created Segment Anything, which performs image segmentation by developing a versatile and promptable model pre-trained on a diverse dataset. By leveraging prompt engineering, the model aims to tackle various segmentation challenges on novel data distributions. You only look once (YOLO) (\cite{yolo}) and Single shot detector (SSD) (\cite{ssd}) proposed by is an object detection algorithm that simultaneously predicts bounding boxes and class probabilities in real-time, enabling efficient and accurate object detection in images and videos. Some other image localization methods include Faster R-CNN (\cite{faster-rcnn}), 
and RetinaNet (\cite{retinanet}). These methods also aim to accurately localize objects in images using different network architectures and techniques. Our approach leverages localization techniques based on multi-modal encoders and gradient mechanisms to improve visual QA via automatic cropping.\footnote{\scriptsize We also experimented with using YOLO and Segment Anything for cropping, but we do not report their results in this paper due to poor performance on the visual QA task.} 


\section{Method}

\subsection{Vision QA Models}


We focus on two BLIP-family models (BLIP and BLIP2), due to their strong performance and speed, and BLIP2's high accuracy on the zero-shot Vision QA task. 
The \textbf{BLIP}~\citep{blip} model consists of an image encoder and pre-trained text-decoder, which enable it to effectively utilize noisy web data by bootstrapping captions. 
Distinctively, BLIP is designed to transfer flexibly to both understanding-based and generation-based tasks, which most existing pre-trained models struggle with. Thus,
BLIP has gained tremendous success across a spectrum of multi-modal downstream tasks such as image-text retrieval (+2.7\% in average R@1), image captioning (+2.8\% in CIDEr), and Visual QA (+1.6\% in VQA-v2 score). 
Its successor, \textbf{BLIP2}~\citep{li2023blip}, 
leverages frozen pre-trained image models and large language models and connects them with a Q-former that maps vision features into text spaces. As a result, BLIP2 reaches state-of-the-art performance on various vision-language tasks while training with significantly fewer parameters. Additionally, the zero-shot ability of BLIP2 increases its robustness and applicability across diverse datasets and real-world scenarios, e.g., BLIP2 can generate personalized image captions based on natural language instructions, without any fine-tuning on a specific dataset and it can perform visual commonsense reasoning by generating a plausible explanation for an image that is not explicitly described in the input prompt.

\begin{figure}[!t]
    \centering
    \includegraphics[width=0.7\textwidth]{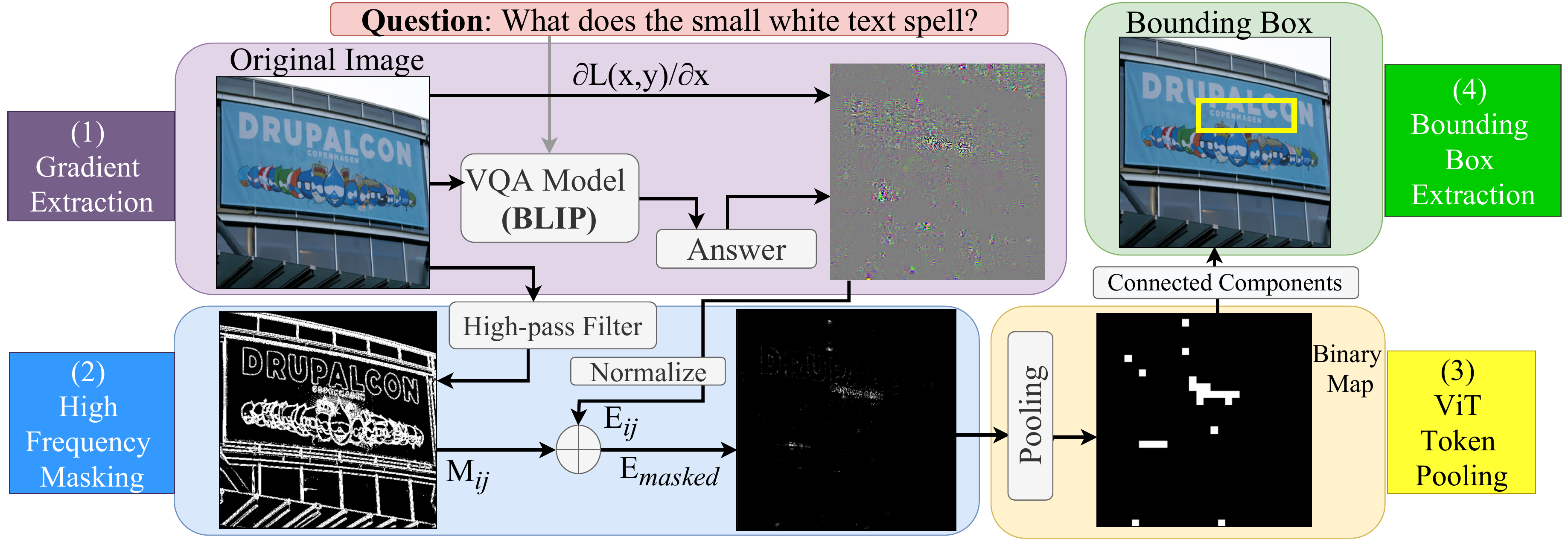}
    \caption{Localization of area of interest using \textbf{gradient-based cropping}.}
    \label{fig: gradient}
\end{figure}

\subsection{Visual Cropping Approaches}

To investigate the utility of cropping, i.e., localizing an area in an image that is most relevant to answer a question, we start with a \texttt{human} cropping procedure. Human cropping is based on examining the image and the questions, followed by drawing the smallest bounding box that encapsulates all the necessary visual information for answering the question. While this strategy enables us to study whether ideal cropping helps the BLIP-family models, it is unrealistic in practice. Therefore, we next present two cropping methods based on BLIP model gradients and multi-modal CLIP encoding, each with its variants. Both localization methods use the entire question as a prompt.


\subsubsection{Gradient-based Cropping (\texttt{grad})}

Due to their high performance, we expect that BLIP-family models are able to recognize the regions of interest within an image using a cross-attention mechanism. Still, their overall design using an image encoder, compels them to process the entire image, without the ability to re-focus solely on the relevant areas. Inspired by this, we devise \texttt{grad}: an innovative cropping pipeline that exploits the parameters of the BLIP visual QA model to pinpoint the areas of interest in the image for addressing the corresponding question. We then use a cropped version of the image centered on these identified regions, thus
ensuring that the model's processing power is directed toward the most relevant portions of the image during the question-answering phase. 
The \texttt{grad} method consists of the following steps (\autoref{fig: gradient}): 1) we \textbf{extract the gradient} of each pixel from the model and normalize it, 2) we apply \textbf{high-frequency masking} to the original image by using a high-frequency filter to produce a mask and applying this mask to the gradient map, 3) we apply percentile\textbf{ token pooling} to the image according to the size of image tokens and make the result a binary image, and 4) we perform \textbf{bounding box extraction} by locating the largest connected region in the binary image and draw the smallest bounding box which covers the whole region. We describe each step in turn.

\textbf{Gradient Extraction}~
Given an image-question pair, the model generates the answer in an auto-regressive way using the language modeling head. We take the answer given by the model and compute the language modeling loss of generating such answer: 
\begin{equation}
\small
L = -\sum_{t=1}^{T} \log p(y_t | y_{<t}, I, Q; \theta)
\end{equation}

where $L$ denotes the language modeling loss, $T$ is the total number of tokens in the answer, $y_t$ is the $t^{th}$ token in the answer, $y_{<t}$ are the tokens before the $t^{th}$ token, $I$ is the image, $Q$ is the question and $\theta$ represents the model parameters. 
We then compute the derivative of the language modeling loss with respect to each pixel. For each pixel, the gradient of $R_{ij}, G_{ij}, and~B_{ij}$ values are computed separately.
We combined the gradients of them for each pixel and emphasize the positive gradients by applying a $ReLU$:

\begin{equation}
\small
E_{ij} = ReLU(\frac{\partial L}{\partial R_{i,j}}) + ReLU(\frac{\partial L}{\partial G_{i,j}}) + ReLU(\frac{\partial L}{\partial B_{i,j}})
\end{equation}

$E_{ij}$ is a combined representation of the gradient of each pixel.
Considering extremely high or low values can distort the scale when normalizing, compressing the majority of values into a narrow range, we discard the highest and lowest $k_{discard}\%$ of gradients before normalization. 
Then we highlight the pixels that have higher $E_{ij}$  by recursively retaining only the values that are greater than their mean:
    
\begin{minipage}{.5\linewidth}
\begin{equation}
\small
\label{formula3}
E^0_{ij} = 
\begin{cases}
E_{ij} & \text{if } P_{k_{discard}} \leq E_{ij} \leq P_{100-k_{discard}} \\
0 & \text{otherwise}
\end{cases}
\end{equation}
\end{minipage}%
\begin{minipage}{.5\linewidth}
\begin{equation}
\small
\label{formula4}
E^{(i+1)}_{ij} = 
\begin{cases}
E^{(i)}_{ij} & \text{if } E^{(i)}_{ij} > \text{mean}(E) \\
0 & \text{otherwise}
\end{cases}
\end{equation}
\end{minipage}

\textbf{High-frequency Masking}
To minimize the influence of irrelevant information (e.g., the difference in brightness of each region), we apply a high-pass Gaussian frequency filtering by subtracting the Gaussian blur from the original image. Then we produce a binary mask image that filters all the pixels larger than the average~\autoref{binary mask}. 

\begin{equation}
\small
\label{binary mask}
M_{ij} = \begin{cases} 
1 & \text{if } H(I^n_{ij}, k, \sigma) > \overline{H(I, k, \sigma)} \\
0 & \text{otherwise}
\end{cases}, H(I^n_{ij}, k, \sigma) = I - G(I^n_{ij}, k, \sigma)
\end{equation}


where $I$ denotes the original image, $G$ is result of the Gaussian blur,
and $K_{size}$ is the kernel size of it, $\sigma$ is the Gaussian kernel standard deviation. 
$H$ denotes the high-frequency pass over the image, $M$ is the final mask. Then the mask is applied to the gradient map: $E_{masked} = E \circ M$. 


\textbf{ViT Token Pooling}
We employ a vision-token-based pooling strategy because image features are primarily extracted via a Vision-transformer (ViT)~\citep{vit}. For each token within image (e.g., a 16$\times$16 patch), we select top $n_{pool}\%$ percentile of values from the corresponding patch of $E_{masked}$ as the representative value for that particular patch. Subsequently, these values are converted into a binary map where values surpassing average are set to $1$, and those below are set to $0$.

\textbf{Bounding Box Extraction}
Finally, we identify the most extensive connected component on the post-pooling binary map and establish the smallest bounding box that encompasses this specific region. To ensure that the bounding box image includes comprehensive content, we expand the bounding box by a factor of 1.5 in each direction.

\subsubsection{CLIP-based Cropping (\texttt{clip-*})}

CLIP~\citep{clip} is a state-of-the-art model for multi-modal encoding, which efficiently learns visual concepts from the text. CLIP uses a transformer-based model to encode both images and text into a shared feature space, which allows for direct alignment between the two modalities and enables zero-shot classification. We devise two cropping variants based on CLIP.

\textbf{Sliding-window Cropping (\texttt{clip-w})} 
divides an image into several square patches $P_{ij}$, each with dimensions of $N*N$.\footnote{\scriptsize \url{https://www.pinecone.io/learn/zero-shot-object-detection-clip}, accessed on May 17th, 2023.} After splitting the image into patches, we use square windows encompassing $w*w$ patches:
\begin{equation}
\small
W_{ij} = I[iN:(i+w)N, jN:(j+w)N]
\end{equation}
where $i \in [0, \left\lceil\frac{W}{N}\right\rceil-w)$ and $j \in [0, \left\lceil\frac{H}{N}\right\rceil-w)$. $W$ and $H$ represent the width and height of the original image, respectively, $\left\lceil\right\rceil$ indicates rounding up, and $w$ denotes window size. Then the CLIP similarity score of a window $W_{ij}$ with respect to the text prompt $(TP)$ is computed, then the window traverses the full image with a stride $s$, resulting in an accumulated similarity score for each patch. The scores are then divided by the number of times the window covered the patch:
\begin{equation}
\small
\bar{A}_{kl} = \frac{\sum_{(i,j): P_{kl} \in W_{ij}} S_{ij}}{\sum_{(i,j): P_{kl} \in W_{ij}}},~
S_{ij} = \text{CLIP-Similarity}(W_{ij}, TP)
\end{equation}
Then the $\bar{A}_{kl}$ is normalized and the high-value is highlighted (see~\autoref{formula4}) to $\hat{A}_{kl}$  and we select all the patches that are higher than a threshold $\theta$. The final bounding box is the smallest rectangle encompassing all selected patches.


\textbf{Recursive Cropping (\texttt{clip-r}})~
Given an image and a text prompt as input, the \texttt{clip-r} algorithm crops the input image from four directions: top, bottom, left, and right, to generate four overlapping cropped images, each cropped by a specific cropping ratio $r$. 
For instance, the height of the top crop will be $r$ times the original height with the width remaining the same as the original bounding box. 
Similarly, the width of the left crop will be $r$ times the original width, and its height will retain. 
The partitioned images are then processed with CLIP~\citep{clip} along with the text prompt to evaluate the semantic similarity between the four cropped images and the prompt. Then the crop with the highest score is selected as the input for the next iteration. This process is repeated for a specified number of iterations, progressively narrowing down the region of interest in the image that semantically aligns best with the text prompt.

\subsection{Dataset Construction}

To investigate whether image cropping improves the ability of BLIP-family models to focus on salient content, we systematically construct three datasets by subsetting existing visual QA data.

\textbf{\texttt{1)} VQA-text} contains reading questions whose answers can be found as symbols, typically a sequence of letters and/or digits in the image. The specific location of symbols is sometimes indicated by a constraint in question, e.g., \textit{what is the brand of the camera placed on the white table?} We build \textit{VQA-text} from the validation set of Text-VQA~\citep{textvqa}. 
To ensure a fair testing ground for a one-time cropping experiment, we perform an additional step that maintains the model's focus on a single textual entity at a time. Namely, we take the Optical Character Recognition (OCR)~\citep{ocr} annotations from the dataset and keep the questions where the OCR result is perfectly aligned with ground truth in a single bounding box. 
After dataset selection, to ensure the inclusion of all relevant textual information, we expand the bounding boxes by a factor of 1.5. 
The resulting VQA-text dataset consists of 500 questions paired with 408 images.

\textbf{\texttt{2)} VQA-hard} collects visual QA model failures that could be hypothetically remedied via cropping. Our collection process starts with the extraction of intersecting failure instances from two distinct visual QA models, the zero-shot visual QA model \textit{BLIP-2-FlanT5XL}~\citep{li2023blip} and the fine-tuned \textit{BLIP-vqav2}~\citep{blip}. 
An example is considered a failure when the answer is not the same as the majority of human annotations.
Then we randomly select 400 samples from the intersection of the model failures, and we reserve the questions which: 1) truly represent a model failure (e.g., leaving out near-synonyms like city and town), and 2) could potentially be rectified by image cropping (i.e., preserving cases with a clear focus on a small number of locations in the image). For the selected examples, we examine their images and questions and draw the smallest bounding box that encapsulates all the necessary visual information for answering the question.
The resulting VQA-hard dataset includes 109 questions with their corresponding images.



\textbf{\texttt{3)} VQA-random:} Due to the significant computational requirements needed to process VQA-v2's validation set, which contains 214K questions, we do not evaluate our cropping model on this entire set. Instead, we randomly select a subset with 1,001 questions and 973 images from VQA-v2 for our evaluation, which we refer to as the VQA-random. This approach allows us to maintain a balance between computational feasibility and the robust evaluation of our methods. 


\section{Experiments}



\textbf{Experimental Setup}~
Our analysis is carried out on the above-mentioned three datasets with zero-shot \textit{BLIP-2-FlanT5XL} and the fine-tuned \textit{BLIP-vqav2}.  We employ our three automatic cropping methods. For VQA-text and VQA-hard, we also utilize existing human croppings. 
In the case of the zero-shot BLIP2 model, images are transformed into the same embedding space as the question text, facilitating easy concatenation prior to input into the language model. We exploit this feature by concatenating embeddings of both the cropped image and the original image to serve as a joint visual input for the language model.


\textbf{Metrics}~
We employ \textit{accuracy (\texttt{acc})} following VQA-v2,\footnote{ \scriptsize https://visualqa.org/evaluation.html} which is a robust metric considering inter-human variability in phrasing. After applying text processing steps to normalize the model answer, the accuracy is defined as:
$\texttt{acc(ans)} = \min(0.3\times n, 1)$, where $n$ denotes the times that the answer ($ans$) appears among the answers from 10 human annotations. In addition to accuracy, we introduce a metric based on \textit{string similarity (\texttt{str-simi})}, which is calculated using the longest common sub-sequence technique,\footnote{ \scriptsize We also tried other string similarity metrics, but they are highly correlated to each other, so we only show results for one of them.} which is able to distinguish similar answers from dissimilar ones. The \texttt{str-simi} metric is particularly relevant in the context of VQA-text, given its inherent nature as a string-recognition task.

\textbf{Implementation Details}~
For \texttt{grad}, we use a $k_{discard}$ of $1$, a kernel size of 5, and $\sigma$ as $0.3 * ((K_{size} - 1) * 0.5 - 1) + 0.8$, which is $1.1$ by default. 
For patch pooling, we use a patch size of $16\times16$ which is the size of ViT tokens. During pooling, we use a top pooling with a $n_{pool}$ of $5$. For \texttt{clip-w}, we use a window size of $6\times6$, a patch size of $16\times16$, a stride of $1$, and a final threshold $\theta$ of 0.5. For \texttt{clip-r}, we use a cropping ratio $r$ of 0.9, and we perform 20 recursive crops for each image.
We use \textit{python 3.8.16, salesforce-lavis 1.0.2, transformers 4.29.1 and torch 2.0.1} for all the BLIP model experiments. Our environment consists of an Intel(R) Xeon(R) Gold 5215 CPU @ 2.50GHz with 40 cores and 256 GB of RAM. Additionally, we utilize NVIDIA RTX A5000 GPUs for our experiments.

\begin{table}[t!]
\centering
\small
\caption{Human cropping and automatic result on our VQA-text, VQA-hard, and VQA-random dataset. We experiment with fine-tuned and zero-shot VQA models. $^+$ denotes the concatenation result of the original image and cropped image. The performance of the best auto-cropping strategy for a model and dataset is marked in bold.}
\begin{tabular}{l|c|rr|rr|rr}
\toprule

\multicolumn{2}{c|}{\textbf{Dataset} \textit{(question count)}} &  \multicolumn{2}{c}{\bf VQA-text (500)} & \multicolumn{2}{c}{\bf VQA-hard (109)} & \multicolumn{2}{c}{\bf VQA-random (1001)} \\ \midrule
\multicolumn{1}{l}{} & \textbf{cropping} & \texttt{acc} & \texttt{str-simi}  & \texttt{acc} & \texttt{str-simi}  & \texttt{acc} & \texttt{str-simi}  \\\midrule
 & w/o cropping & 27.34 & 60.47 & 44.36 & 68.14 & 83.53 & 94.56 \\
 & \hc & 30.68 & 65.29 & 55.14 & 73.21 & - & - \\
 & \grad & \bf 24.80 & \bf 59.58 & 48.90 & 67.32 & 67.61 & 83.87 \\
 & \sac & 22.80 & 55.96 & \bf 49.63 & \bf 70.95 & \bf 79.48 & \bf 92.00 \\
\multirow{-5}{*}{Fine-tuned} & \rac & 19.86 & 53.04 & 47.34 & 69.65 & 71.06 & 86.31 \\\midrule
 & w/o cropping & 27.76 & 57.52 & 33.60 & 55.55 & 60.98 & 79.06 \\
 & \hc & 57.26 & 83.04 & 41.83 & 64.59 & - & - \\
 & \hc$^{+}$ & 58.52 & 82.81 & 41.93 & 61.69 & - & - \\
 & \grad & 37.12 & 66.28 & 40.64 & 63.74 & 59.83 & 79.08 \\
 & \grad$^{+}$ & \bf 43.04 & \bf 70.39 & \bf 41.47 & \bf 63.32 & \bf 65.57 & \bf 82.70 \\
 & \sac & 27.62 & 56.32 & 35.41 & 57.29 & 60.59 & 78.57 \\
 & \sac$^{+}$ & 32.54 & 59.64 & 34.86 & 55.17 & 61.94 & 79.29 \\
 & \rac & 25.80 & 54.61 & 36.15 & 57.26 & 60.71 & 79.14 \\
\multirow{-9}{*}{Zero-shot} & \rac$^{+}$ & 32.70 & 61.13 & 35.96 & 57.67 & 63.44 & 80.29 \\
\bottomrule
\end{tabular}

\label{tab:main_result}
\end{table}

\section{Results}

\subsection{Effectiveness of Cropping Strategies}

\textbf{Human cropping helps VQA performance, especially for zero-shot models.} \autoref{tab:main_result} shows the performance of the BLIP fine-tuned and the BLIP2 zero-shot model with different cropping methods compared to the original baseline. We note that providing a human-cropped bounding box enhances the model performance significantly for both VQA-text and VQA-hard. On the VQA-text dataset, the performance increases by 3 absolute points for the fine-tuned model and by 30\% (206\% in relative terms) for the zero-shot model. Curiously, with this boost in its performance, the zero-shot model's performance on reading text from images becomes superior to that of the fine-tuned model. We attribute this to the overfitting of the fine-tuned model to the VQA dataset, making it unable to generalize to cropped images that are comparatively easier but uncommon in the training data. Cropping is also highly beneficial for the VQA-hard subset, where the models' performance with human cropping grows by 11\% for both the fine-tuned model and the zero-shot model. We do not have human cropping results on the VQA-random set.

\textbf{Automatic cropping can partially approximate human cropping.} As human cropping cannot be expected to be realistically available for novel images, we investigate to what extent our automatic methods can approximate human cropping results. Our results (\autoref{tab:main_result}) show that automatic cropping is able to approximate human cropping on the VQA-hard dataset. Curiously, automatic cropping is less effective on the VQA-text data, where it harms performance for the fine-tuned model and performs between the baseline and human cropping on the zero-shot model. This shows that detecting the correct positioning of text is more challenging for automatic cropping methods than the wide set of questions in the VQA-hard set. 
We observe analogous results on the VQA-random dataset, where again automatic cropping methods help the zero-shot performance but harm the fine-tuned model performance.
We conclude that automatic cropping methods can often mimic human cropping performance, yet, certain question types may require specialized cropping methods.



\subsection{Factors Impacting Cropping Performance}
\label{5.2}


\textbf{Gradient cropping works better with the zero-shot model, concatenation provides a further boost.} The results in \autoref{tab:main_result} show that the \texttt{grad} method performs better on average than the \texttt{clip} models. 
This performance difference is especially clear for the zero-shot model, where \texttt{grad} performs better across all three datasets with a substantial margin. The performance difference for the fine-tuned model is less pronounced, with the \texttt{clip-w} model performing slightly better on average, and performing especially well on the VQA-random set. Among the two \texttt{clip} variants, the \texttt{clip-w} method performs consistently better on the fine-tuned model, and the \texttt{clip-r} model performs better on the zero-shot model. We hypothesize that this is because using a static sliding window aligns better with the VQA-v2 data images on which BLIP has been fine-tuned, whereas the zero-shot BLIP2 model is more flexible and able to benefit from dynamic window sizes as it overfits less to the VQA-v2 images. Finally, we note that the impact of the concatenated model (denoted by `+' in the table) is consistently positive across different datasets and cropping methods. Notably, the integration of the gradient-cropped image and the original image improves the zero-shot performance on the VQA-random set by 4.59 percentage points, which indicates that the zero-shot model is able to flexibly weigh the fine detail and the original image at inference time.

\begin{table}[t!]
\centering
\small
\caption{Granular performance of different sizes of bounding boxes, where the ratio is the bounding box's area divided by the whole image's area. We mark the best result per model and dataset in bold. }
\begin{tabular}{l|c|rr|rr|rr}
\toprule
\multicolumn{2}{c}{\textbf{bbox\_ratio}} \textit{(question count)} & \multicolumn{2}{|c}{\textbf{$<$ 0.005} \textit{(90)}}&\multicolumn{2}{|c}{\textbf{$<$ 0.05} \textit{(410)}}  &\multicolumn{2}{|c}{\textbf{$>$ 0.05} \textit{(90)}} \\ \midrule
\textbf{Model} & \textbf{Cropping Method} & \texttt{acc} & \texttt{str-simi} & \texttt{acc} & \texttt{str-simi}  & \texttt{acc} & \texttt{str-simi} \\ \midrule
\multirow{2}{*}{Fine-tuned} 
& w/o cropping & 22.11 & 56.30 & 26.17 & 59.92 & \bf 34.22 & 65.81 \\ 
& \hc  & \bf 28.44 & \bf 61.17 & \bf 29.98 & \bf 64.92  & 33.89 & \bf 66.99\\
\midrule
\multirow{3}{*}{Zero-shot}  
& w/o cropping & 11.33 & 45.69  & 22.78 & 52.58 & 56.22 & 84.32 \\ 
& \hc & 64.22 & \bf 85.21 & 57.78 & \bf 82.93 & 54.89 & 83.56 \\ 
& \hc$^{+}$ & \bf 66.11 &  84.46 & \bf 58.12 & 82.13 & \bf 60.33 & \bf 85.93 \\ 
\bottomrule
\end{tabular}

\label{tab:bbox_size}
\end{table}

\textbf{Cropping helps more when bounding boxes are smaller.} 
We investigate the impact of varying cropping result sizes on our VQA-text dataset. 
\autoref{tab:bbox_size} shows that human cropping helps for different bounding box sizes, yet, its impact is most significant for questions with focused bounding boxes. This method improves the baseline zero-shot model performance by nearly 600\% on the smallest bounding box size questions. The questions which have larger bounding boxes resemble the original image closer, which makes the impact of cropping intuitively less significant, and sometimes even negative, on these questions. Meanwhile, concatenating the cropped and the original image has a positive impact across all three datasets, which shows that the BLIP2 model can consistently combine images with different granularities.

\begin{figure}[!t]
    \centering
    \includegraphics[width=\textwidth]{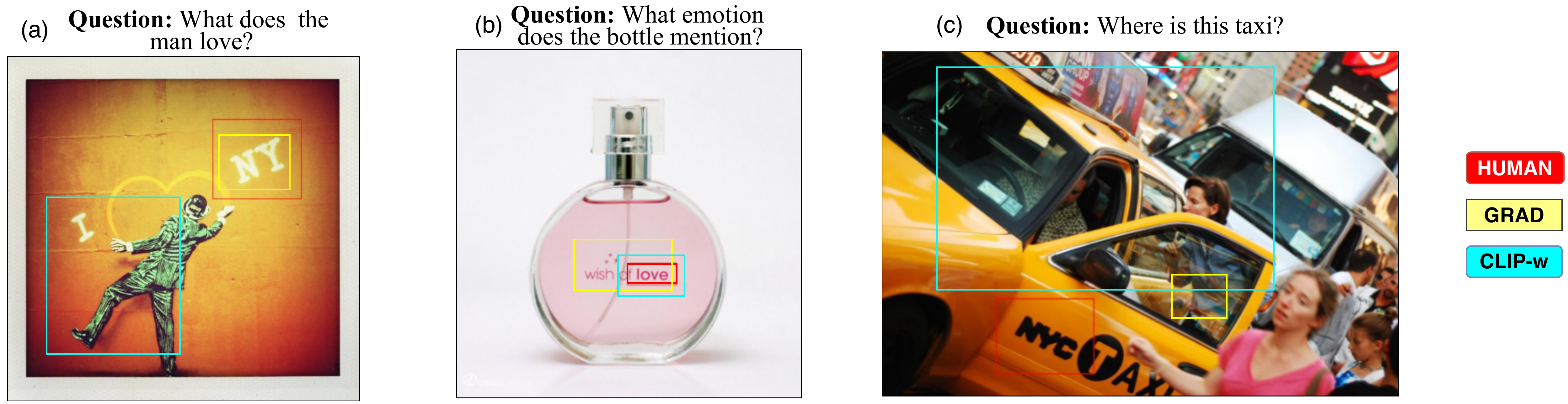}
    \caption{Three cases of bounding boxes given by human and automatic cropping methods.}
    \label{fig:case_study}
\end{figure}

\subsection{Case Study}

We study the predictions of the two automatic cropping methods that use a strategy of picking a bounding box from a patch map: \grad~and \sac, together with human cropping. 
\autoref{fig:case_study} shows three examples. In the left image, humans and \texttt{grad} answer correctly by selecting the correct region (\textit{NY}) precisely. Interestingly, here the CLIP model crops the image around the man, which is the only noun in the question (\textit{what does the man love}). In the middle image, the cropping result of \texttt{clip-w} is close to the one of the human. The image on the right has wrong cropping by both automatic methods, indicating a more complex localization task.


Overall, we observe that \texttt{clip-w} consistently crops a larger area of the image, which is primarily driven by the question keywords. The gradient-based model aims to capture the area of interest more precisely. While \texttt{grad} is able to benefit the visual QA model more, especially in scenarios where the answer was more straightforward and easily captured, we find that the more flexible area selection by CLIP-based models proved advantageous in cases where the answer was distributed across a broader region in the image. These observations are intuitive, as the gradients from the visual QA model are more granular between the pixels while the CLIP scores are computed in a more coarse manner. These insights highlight the varying strengths and considerations associated with automated cropping models in different contexts.

\section{Conclusions and Outlook}


This paper focused on improving the accuracy of state-of-the-art visual QA models on fine-detail questions using visual cropping. We defined three controlled benchmarks and experimented with novel cropping approaches based on the visual QA model's gradient and multi-modal encoding. Our experiments showed that the family of the BLIP model benefits significantly from human cropping, and automated cropping approaches can largely approximate these benefits. A deeper dive into our findings indicates that the performance enhancement is more pronounced in zero-shot models than in fine-tuned models and more salient with smaller bounding boxes than larger ones. While cropping enhancement helps with fine-detail questions, it is also robust on the overall VQA-v2 task. These insights provide a promising direction for improving visual QA models, especially on fine-detail questions, and highlight the importance of further research on this topic. 

Despite our advancements, our experiments exhibit several limitations that open up opportunities for future research.
(1) The cropping process may compromise visual QA performance by eliminating valuable context information from the full image. We attempted to address this issue by concatenating the original and cropped images for visual input. However, this approach deviates from the pre-training regimen of the visual QA model, which might cause mismatches between the learning objectives. 
Future work could explore more sophisticated methods for combining predictions from the cropped and original images, such as training a decision-making model or leveraging large language models to ensemble multiple sources of information~\cite{hugginggpt}.
(2) Our approach is time-consuming. We refrained from testing our method on the entire VQA dataset and chose a random subset instead due to the expensive cropping procedures. For instance, applying a single CLIP sliding window auto-cropping method to our VQA-random dataset with 1001 questions takes approximately 1.5 hours. The recursive strategy is twice as fast as the sliding window approach, yet, it performs relatively worse on average. Therefore, future steps should focus on enhancing the speed and precision of cropping techniques. 
(3) As all our test datasets are subsets of VQA-v2, we see extending our experiments to additional benchmarks as crucial. Real-world images can have a variety of complex factors, e.g., they may be blurred, with low resolution, have less brightness, or have low contrast. To enable our cropping methods to benefit these use cases, we plan to enrich them with image super-resolution and image enhancement methods.


\bibliography{cite}
\bibliographystyle{nips}



\newpage
\appendix
\onecolumn
\section{Results using segmentation and object detection models}

In addition to the auto-cropping methods, we also try object detection and image segmentation methodologies to extract the relevant area of interest corresponding to a given question. To achieve this, we utilized two state-of-the-art models: Segment Anything Model (SAM)~\citep{kirillov2023segment} and YOLO-v8\footnote{https://github.com/ultralytics/ultralytics}~\citep{yolo}, which specialize in segmentation-based and object detection with localization approaches. 

\begin{table}[h!]
\centering
\small
\caption{Complete results including YOLO and SAM model.}
\begin{tabular}{l|c|rr|rr|rr}
\toprule

\multicolumn{2}{c|}{\textbf{Dataset} \textit{(question count)}} &  \multicolumn{2}{c}{\bf VQA-text (500)} & \multicolumn{2}{c}{\bf VQA-hard (109)} & \multicolumn{2}{c}{\bf VQA-random (1001)} \\ \midrule
\multicolumn{1}{l}{} & \textbf{cropping} & \texttt{acc} & \texttt{str-simi}  & \texttt{acc} & \texttt{str-simi}  & \texttt{acc} & \texttt{str-simi}  \\\midrule
 & w/o cropping & 27.34 & 60.47 & 44.36 & 68.14 & 83.53 & 94.56 \\
 & \hc & 30.68 & 65.29 & 55.14 & 73.21 & - & - \\
 & \grad & \bf 24.80 & \bf 59.58 & 48.90 & 67.32 & 67.61 & 83.87 \\
 & \sac & 22.80 & 55.96 & \bf 49.63 & \bf 70.95 & \bf 79.48 & \bf 92.00 \\
 & \rac & 19.86 & 53.04 & 47.34 & 69.65 & 71.06 & 86.31 \\
 & \texttt{yolo}&17.04&49.04&42.94&65.21&68.18&83.45 \\
\multirow{-7}{*}{Fine-tuned} & \texttt{sam}&15.58&48.29&48.72&70.04&65.24&81.83 \\\midrule
 & w/o cropping & 27.76 & 57.52 & 33.60 & 55.55 & 60.98 & 79.06 \\
 & \hc & 57.26 & 83.04 & 41.83 & 64.59 & - & - \\
 & \hc$^{+}$ & 58.52 & 82.81 & 41.93 & 61.69 & - & - \\
 & \grad & 37.12 & 66.28 & 40.64 & 63.74 & 59.83 & 79.08 \\
 & \grad$^{+}$ & \bf 43.04 & \bf 70.39 & \bf 41.47 & \bf 63.32 & \bf 65.57 & \bf 82.70 \\
 & \sac & 27.62 & 56.32 & 35.41 & 57.29 & 60.59 & 78.57 \\
 & \sac$^{+}$ & 32.54 & 59.64 & 34.86 & 55.17 & 61.94 & 79.29 \\
 & \rac & 25.80 & 54.61 & 36.15 & 57.26 & 60.71 & 79.14 \\
 & \rac$^{+}$ & 32.70 & 61.13 & 35.96 & 57.67 & 63.44 & 80.29 \\
 & \texttt{yolo}&19.46&49.10&34.22&56.88&58.10&77.05 \\
 & \texttt{yolo}$^{+}$&29.58&58.74&35.87&58.17&63.70&80.50 \\
 & \texttt{sam}&22.96&53.11&38.17&58.07&57.33&76.38 \\
\multirow{-12}{*}{Zero-shot} & \texttt{sam}$^{+}$&35.14&63.24&36.33&54.04&64.27&80.90 \\
\bottomrule
\end{tabular}

\label{tab:sam yolo}
\end{table}

\subsection{Method description}
We initially fed the input image to the SAM model, generating multiple segmentation masks representing different regions within the image. Subsequently, the smallest bounding box encompassing each segmentation mask was individually passed through the CLIP model to identify the mask that exhibited the highest similarity with the provided question. The segmentation mask exhibiting the highest similarity was then forwarded to the subsequent BLIP model for further visual QA pipeline.
Regarding the YOLO methodology, we followed a similar procedure as with SAM. After inputting the image into the YOLO model, it provided a set of bounding boxes representing detected objects within the image. Each bounding box was subsequently processed through the CLIP model to determine the bounding box that demonstrated the highest similarity with the given question. The bounding box associated with the object exhibiting the highest similarity was then forwarded to the subsequent BLIP model for the visual QA task.

\subsection{Results}
Table \ref{tab:sam yolo} shows the overall comparison of all the cropping methods used in our experimentation. The proposed \grad~model demonstrates a remarkable similarity to the human cropping process, thereby surpassing alternative cropping methodologies in terms of performance and effectiveness. Also, as the SAM and YOLO methods generally performed worse, we left them out of the main paper.

Looking at concrete predictions, the SAM model provides a more accurate representation of the area of interest compared to YOLO. This distinction arises from the methodology employed by SAM, which generates segmentation masks covering the entirety of the image. Consequently, we can individually pass each image segment to the CLIP model, enabling CLIP to perceive the entire image comprehensively.

\begin{figure}
    \centering
    
    \begin{subfigure}{0.32\textwidth}
        \centering
        \includegraphics[width=\linewidth]{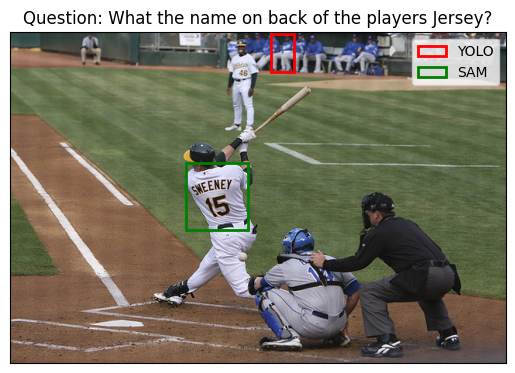}
        \caption{}
        \label{fig:sub1}
    \end{subfigure}
    \hfill
    \begin{subfigure}{0.32\textwidth}
        \centering
        \includegraphics[width=\linewidth]{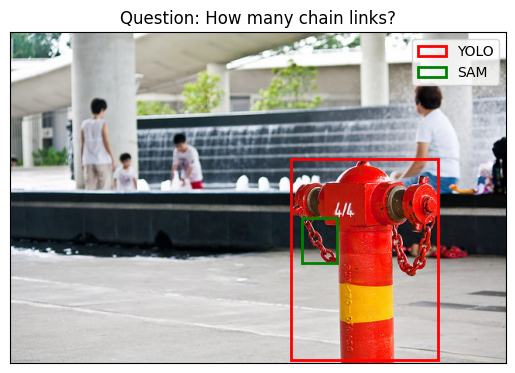}
        \caption{}
        \label{fig:sub2}
    \end{subfigure}
    \hfill
    \begin{subfigure}{0.32\textwidth}
        \centering
        \includegraphics[width=\linewidth]{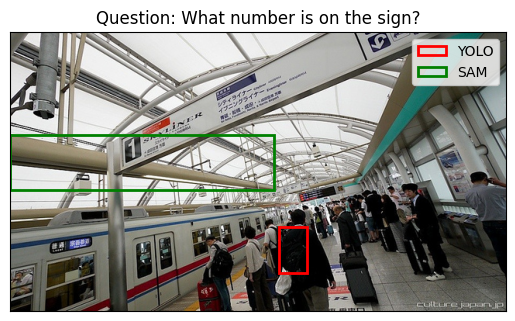}
        \caption{}
        \label{fig:sub3}
    \end{subfigure}
    
    \vspace{0.1cm}
    
    \begin{subfigure}{0.32\textwidth}
        \centering
        \includegraphics[width=\linewidth]{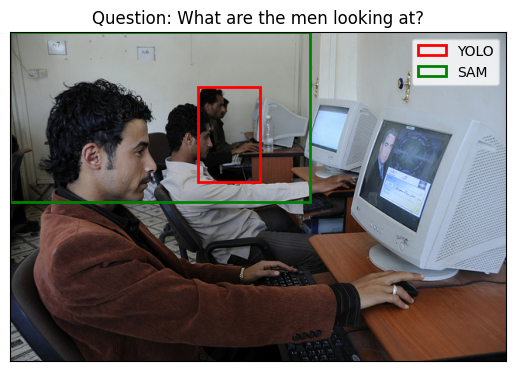}
        \caption{}
        \label{fig:sub4}
    \end{subfigure}
    \hfill
    \begin{subfigure}{0.32\textwidth}
        \centering
        \includegraphics[width=\linewidth]{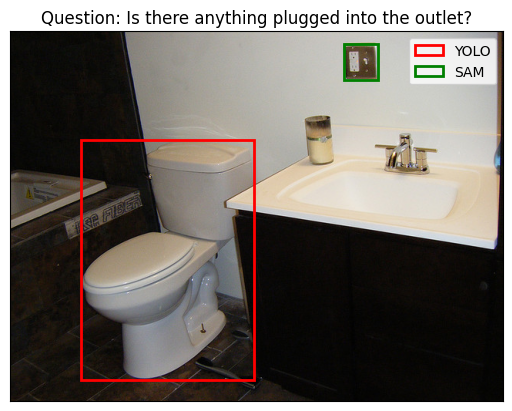}
        \caption{}
        \label{fig:sub5}
    \end{subfigure}
    \hfill
    \begin{subfigure}{0.32\textwidth}
        \centering
        \includegraphics[width=\linewidth]{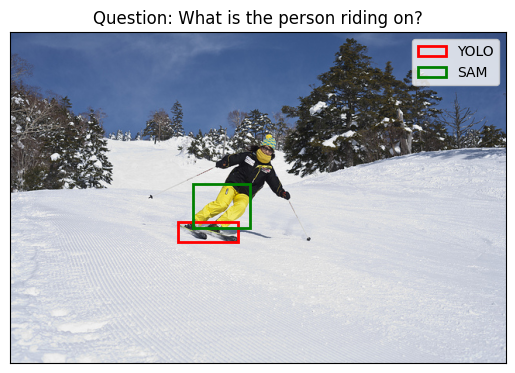}
        \caption{}
        \label{fig:sub6}
    \end{subfigure}
    
    \vspace{0.1cm}
    
    \begin{subfigure}{0.32\textwidth}
        \centering
        \includegraphics[width=\linewidth]{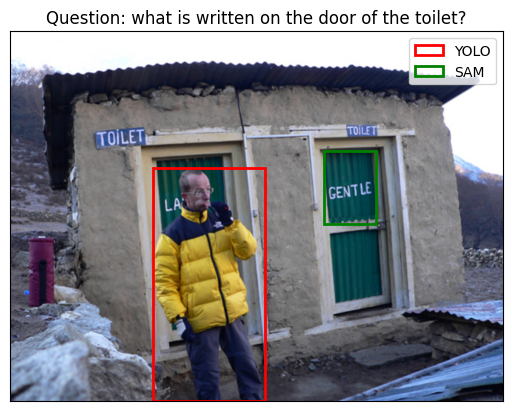}
        \caption{}
        \label{fig:sub7}
    \end{subfigure}
    \hfill
    \begin{subfigure}{0.32\textwidth}
        \centering
        \includegraphics[width=\linewidth]{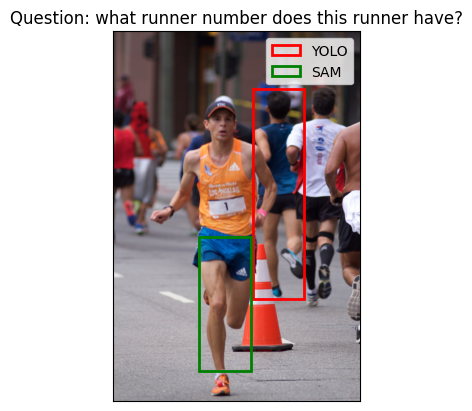}
        \caption{}
        \label{fig:sub8}
    \end{subfigure}
    \hfill
    \begin{subfigure}{0.32\textwidth}
        \centering
        \includegraphics[width=\linewidth]{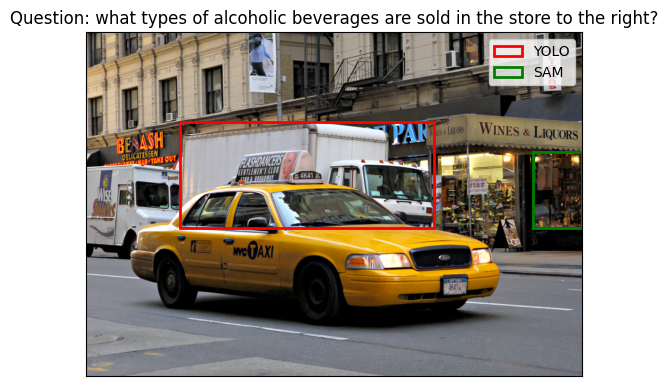}
        \caption{}
        \label{fig:sub9}
    \end{subfigure}
    
    \caption{Examples of bounding box creation using SAM and YOLO.}
    \label{fig:sam_yolo_examples}
\end{figure}

In contrast, YOLO selects objects solely from a predefined list of 80 classes. As a result, there are instances where the detected object may not be directly relevant to the given question. To address this limitation, we employ a strategy to select the bounding boxes that exhibit the closest proximity to the question. However, due to YOLO's class-based approach, it is not an optimal choice for accurately identifying the area of interest. Figure \ref{fig:sam_yolo_examples} shows the bounding box created by SAM and YOLO. In analyzing the bounding boxes, it is evident that SAM presents a superior approach for visual QA due to its ability to localize image regions that hold greater relevance to the posed question. Conversely, being constrained to a limited set of objects, YOLO fails to capture crucial areas within the image that may contain the answer, thus leading to potential information loss.

\begin{table}[ht]
\begin{minipage}[b]{0.45\linewidth}
\centering
\small
\caption{Intersection over Union (IoU in \%) results between the bounding boxes generated by different cropping methods and the human-cropping which we consider as ground-truth.}
\label{tab:iou_results}
\begin{tabular}{lcc}
\toprule
\textbf{Model} & \textbf{VQA-hard} & \textbf{VQA-text} \\ \midrule
\grad   &  \textbf{25.84}  &  \textbf{16.15}    \\ 
\sac   & 23.31   & 5.86       \\ 
\rac &  22.57  &   5.76     \\ 
\texttt{SAM}     &  21.18  &   6.26     \\ 
\texttt{YOLO}    &  20.42  &     3.54   \\ \bottomrule
\end{tabular}
\end{minipage}
\hfill
\begin{minipage}[b]{0.45\linewidth}
\centering
\small
\caption{Average inference time (in seconds/image) for different cropping methods.}
\label{tab:inference_time}
\begin{tabular}{lc}
\toprule
\textbf{Model} & \textbf{Inference time (seconds per image)}  \\ \midrule
\grad   &  0.713      \\ 
\sac   & 5.151      \\ 
\rac &  1.320     \\ 
\texttt{SAM}     &  6.858    \\ 
\texttt{YOLO}    &  \textbf{0.461} \\ \bottomrule
\end{tabular}
\end{minipage}
\end{table}

\section{Comparison of bounding box and inference time of each model}

\subsection{Bounding box comparison}

In the results summarized in \autoref{tab:iou_results}, the \grad~model outperforms the others significantly on both the VQA-hard and VQA-text tasks. This model achieved an IoU of 25.84\% and 16.15\%, respectively. In the context of the VQA-text dataset, our proposed \grad~cropping technique exhibited remarkable proficiency in localizing the specific region within the image where the relevant text was located. In contrast, alternative cropping methodologies consistently faltered, leading to a significantly diminished IoU score across multiple instances.

\subsection{Inference time comparison}
In \autoref{tab:inference_time}, we present a comparative analysis of the inference time associated with various cropping methods. The inference time is calculated as the average time, in seconds, required to process each input image. This calculation is based on an evaluation of 1001 images from the VQA-random dataset, utilizing a single NVIDIA RTX A5000 GPU.

Our observations indicate that the YOLO model, employed for bounding box generation, is the most time-efficient. However, it is essential to note that this efficiency is somewhat compromised by the model's limited class range. On the other hand, the \grad~method, with an inference time of 0.713 seconds, offers a comparable speed to the YOLO model while maintaining superior performance. This balance between speed and performance makes the \grad~a compelling method for bounding box generation.

\section{Ablation study and external analysis}
In this section, we study the effects of our cropping methods' critical components and hyper-parameters by ablation study and external analysis. The experiments in this section provide a better understanding of the performance and properties of different cropping methods.

\subsection{Ablation study of \texttt{grad} components}


In~\autoref{tab:ablation grad}, we apply the ablation study on two critical elements that form the backbone of gradient-based cropping, namely, high-pass filtering (\autoref{binary mask}) and gradient value highlighting (\autoref{formula3} and \autoref{formula4}). Our findings underscore that both these components play pivotal roles in the success of our gradient-based cropping approach.
High-pass filtering comes into play as a crucial function that emphasizes the details in an image, compared to color and brightness. This allows the resulting mask to remove high gradient regions that contain little to no details since it is often such details that determine the precise location a model must focus on. Furthermore, the process of gradient value highlighting serves to temper the extreme values (removing outliers), thereby facilitating a more balanced normalization. It further accentuates regions possessing strong gradients.

\begin{table}[h!]
\centering
\small
\caption{Ablation studies to measure the importance of gradient value highlighting and high-pass filtering for the \grad~method.}
\begin{tabular}{l|c|rr|rr|rr}
\toprule

\multicolumn{2}{c|}{\textbf{Dataset}} &  \multicolumn{2}{c}{\bf VQA-text} & \multicolumn{2}{c}{\bf VQA-hard } & \multicolumn{2}{c}{\bf VQA-random} \\ \midrule
\multicolumn{1}{l}{} & \textbf{metric} & \texttt{acc} & \texttt{str-simi}  & \texttt{acc} & \texttt{str-simi}  & \texttt{acc} & \texttt{str-simi}
\\\midrule
 & w/o cropping & 27.76 & 57.52 & 33.60 & 55.55 & 60.98 & 79.06 \\
 & \grad$^{+}$ & \textbf{41.47} & \textbf{63.32} & \textbf{43.04} & \textbf{70.39} & \textbf{65.57} & \textbf{82.70 }\\
 & w/o grad value highlighting& 35.32 & 54.64 & 39.98 & 67.27 & 65.28 & 81.96 \\
 \multirow{-4}{*}{Zero-shot} & w/o high-pass filtering & 34.40 & 54.46 & 34.34 & 61.42 & 64.22 & 81.48\\

\bottomrule
\end{tabular}

\label{tab:ablation grad}
\end{table}

However, the influence of both high-pass filtering and gradient value highlighting seems to lessen on the VQA-random dataset. The accuracy and structural similarity (str-simi) values remain relatively steady despite the omission of either factor.

\begin{table}[h]
\small
\centering
\caption{Experimental results of two clip-based models on changing the granularity of string-image similarity.}
\label{tab:external study clip}
\begin{tabular}{ccc|ccc}
\toprule
\multicolumn{3}{c}{\sac} & \multicolumn{3}{c}{\rac} \\\midrule
iter-10, ratio-0.2 & \multicolumn{1}{c}{iter-20, ratio-0.1} & iter-40, ratio-0.05 & large-patch & \multicolumn{1}{l}{medium-patch} & \multicolumn{1}{l}{small-patch} \\
\multicolumn{1}{c}{34.59} & 34.86 & 36.33 & \multicolumn{1}{c}{35.32} & 35.96 & 36.24 \\
\multicolumn{1}{c}{54.86} & 55.17 & 56.7 & \multicolumn{1}{c}{55.55} & 57.67 & 56.45\\\bottomrule
\end{tabular}
\end{table}

\subsection{Impact of the granularity of similarity computation}

To study whether more granular image-text similarity computation improves the cropping performance, we adjust the hyperparameters of \rac~and \sac. Specifically, we modify the patch size and window size of the \sac~model, while ensuring that each window contains the same number of pixels. We introduce three variations for the patch size: 8, 16, and 32, with corresponding window sizes of 12, 6, and 3, respectively. For the \rac~model, we alter the number of iteration times to 10, 20, and 40, and adjust the cropping ratios to 0.2, 0.1, and 0.05, respectively. The adjustments are designed to maintain approximately the same final cropping size.

As presented in Table~\ref{tab:external study clip}, our observations suggest that increasing the granularity of the two clip-based cropping methods results in a modest performance enhancement. The most substantial increase we observe is 1.74\%. This finding indicates that the model's granularity does not play a decisive role in its performance.

\section{More examples}

In Figure \ref{fig: gradient}, we provide more examples of different models' successes and failures in predicting the correct answer for the given question. We also provide the bounding boxes created by each of the three models (\grad,~\sac,~ and \rac).

\begin{figure}[!h]
    \centering
    \includegraphics[width=0.9\textwidth]{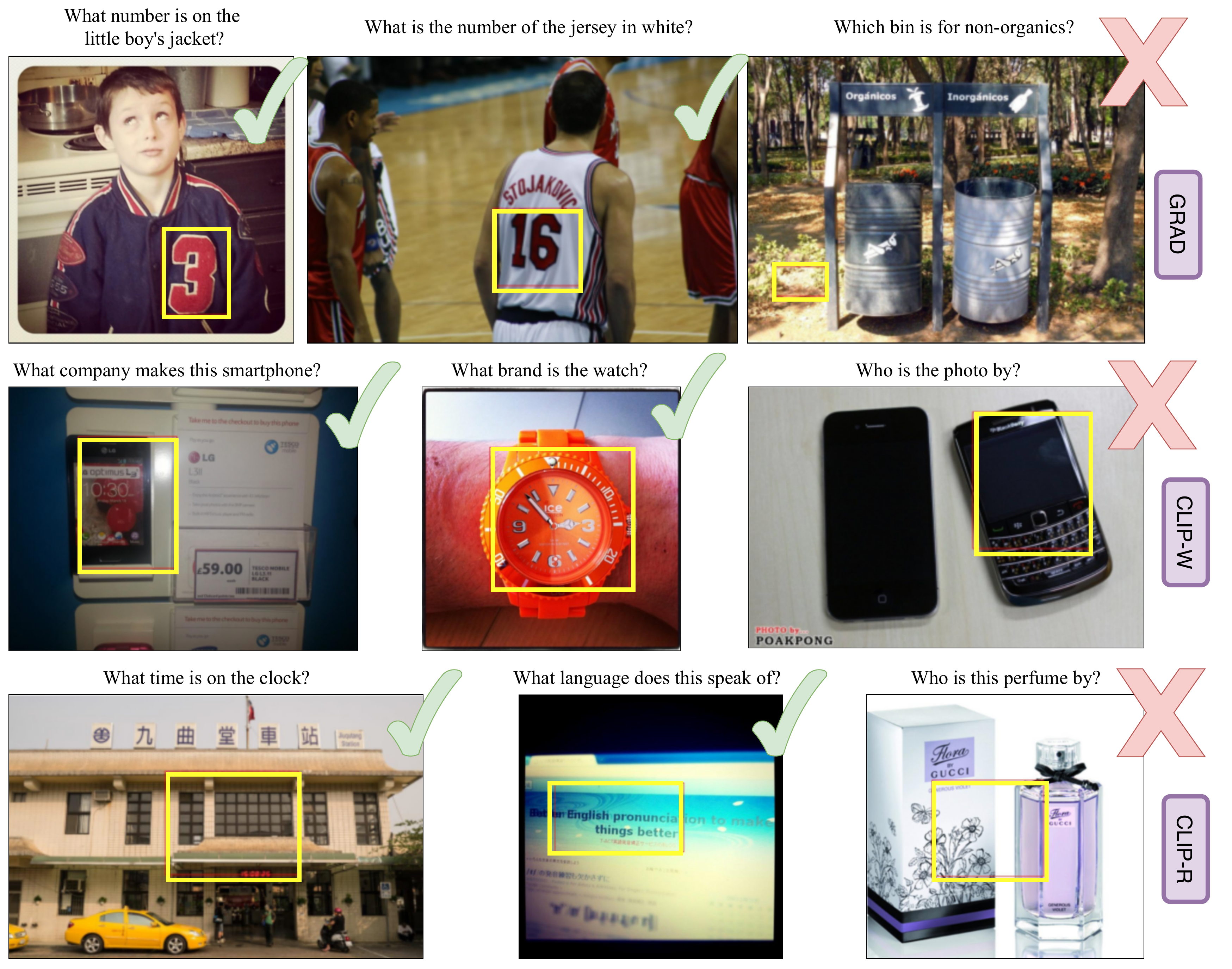}
    \caption{Success ($\checkmark$) and Failure ($\times$) of the used cropping techniques in predicting the correct answer to the question.}
    \label{fig: gradient}
\end{figure}


\end{document}